\documentclass{article}
\usepackage{spconf,amsmath,amssymb,graphicx,booktabs,url}

\title{COMPILING AUTOREGRESSIVE LLM CONTROL INTO POLICY-EXACT\\ PARALLEL INFERENCE FOR HIERARCHICAL RAG}

\makeatletter
\gdef\@name{{\em Zehao Lin$^{1}$, Qianfeng Shen$^{2}$, Xixuan Hao$^{1,3}$,
  Shaobo Cui$^{4}$, Chunyu Li$^{1}$, Kai Chen$^{1}$,}\\
  {\em Bo Tang$^{1}$, Zhenjun Wang$^{2}$, Jin Yan$^{2}$,
  Zhiyu Li$^{1}$\sthanks{Corresponding author.}, Feiyu Xiong$^{1}$}\\}
\makeatother
\address{$^{1}$MemTensor (Shanghai) Technology Co., Ltd.
  \quad $^{2}$China Haisum Engineering Co., Ltd.\\
  $^{3}$The Hong Kong University of Science and Technology (Guangzhou)
  \ $^{4}$Shanghai Jiao Tong University\\
  {\small\tt \{linzh,chenk,licy,tangb,lizy,xiongfy\}@memtensor.cn,
  shaobo.cui@sjtu.edu.cn}\\
  {\small\tt \{shenqf,wangzhenjun,yanjin\}@haisum.com,
  xhao390@connect.hkust-gz.edu.cn}}

\begin{document}
\ninept
\maketitle

\begin{abstract}
% Hierarchical RAG places an LLM in a control loop: at each frontier of a document tree it commits to $b$ of $k$ sections under an input-dependent action grammar. Deployed systems execute it by decoding text: either failure-prone free-form JSON or reliable but strictly serial grammar-constrained generation. We show that the controller need not be decoded at all: the bounded horizon makes the reachable future a shallow action-prefix tree, which we \emph{compile} --- one root prefill plus one tree-causal probe evaluate every prefix-conditioned action distribution with serial decoding's tokens, positions, and causal ancestry. A prefix-state lemma makes compiled control \emph{policy-exact in exact arithmetic}: the same policy with a two-forward critical path and zero output tokens. One further probe pass certifies the exact order-marginalized MAP set.
% Across four document-disjoint splits and two scales on the primary MetaX device, compiled control matches the serial policy on $84$--$89\%$ of questions --- every instrumented divergence a floating-point near-tie --- cutting control latency by $35$--$47\%$ versus token-wise serial decoding and $6$--$7\%$ versus an idealized chunked engine; an H800 repetition preserves the ordering.
% I think current version lacks motivation and feels ambiguous. So, I wrote my version -- Shaobo
Large language models increasingly act as controllers in structured
decision pipelines, yet their decisions are still executed by
autoregressive text decoding. Hierarchical RAG is one such setting: at
each frontier of a document tree, an LLM commits to $b$ of $k$ sections
under an input-dependent action grammar, executed as either
failure-prone free-form JSON or reliable but strictly serial
grammar-constrained generation. We show that this serialization is
unnecessary for bounded-horizon control: the reachable future forms a
shallow action-prefix tree, which we \emph{compile} --- one root prefill
plus one tree-causal probe evaluate every prefix-conditioned action
distribution with serial decoding's tokens, positions, and causal
ancestry. A prefix-state lemma makes compiled control
\emph{policy-exact in exact arithmetic}: the same policy with a
two-forward critical path and zero output tokens. One further probe
pass certifies the exact order-marginalized MAP set. Across four
document-disjoint splits and two scales on the primary MetaX device,
compiled control matches the serial policy on $84$--$89\%$ of questions
--- every instrumented divergence a floating-point near-tie --- and cuts
control latency by $35$--$47\%$ versus token-wise serial decoding and
$6$--$7\%$ versus an idealized chunked engine; an H800 repetition
preserves the ordering.
\end{abstract}


\begin{figure*}[t]
\centering
\includegraphics[width=0.94\textwidth]{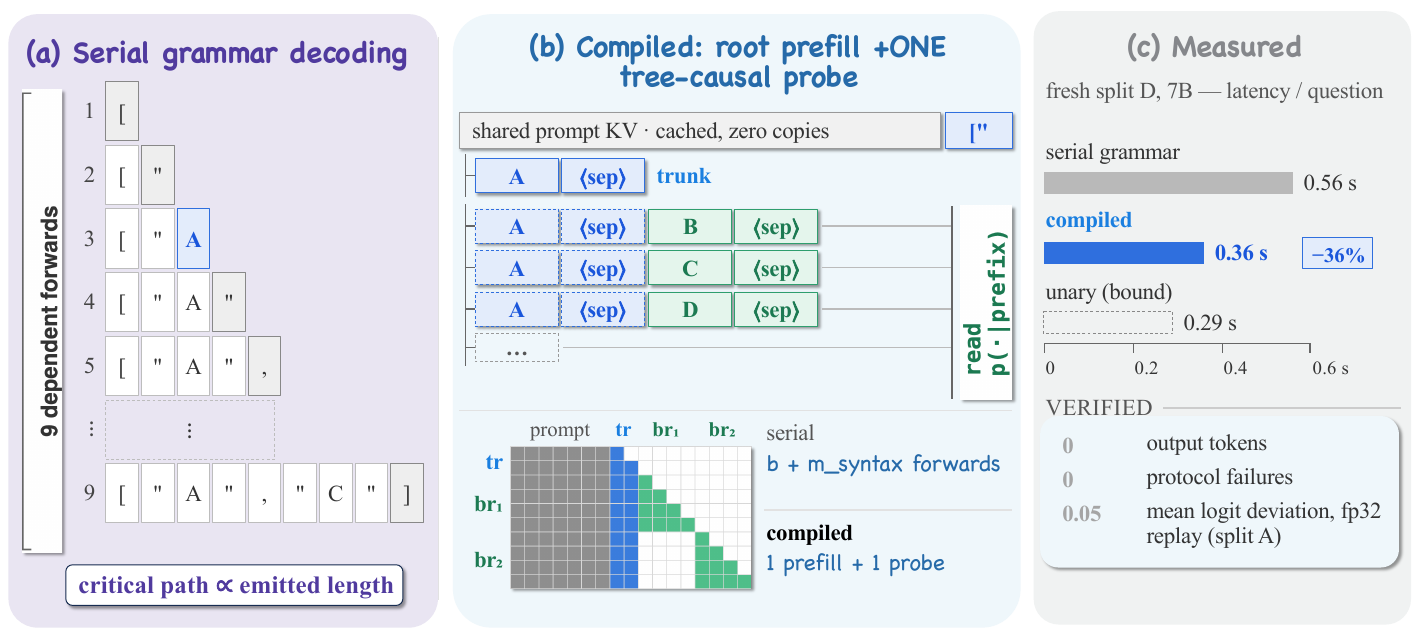}
\caption{Compile, don't decode. (a)~Serial grammar decoding spends one
dependent forward per emitted token ($b+m_{\mathrm{syn}}\approx 9$).
(b)~Compilation: one root prefill, then a single tree-causal probe
evaluates the trunk and every candidate branch under a block-causal mask
--- each row sees only the cached prompt and its own ancestors, so its
logits equal the serial controller's in exact arithmetic (Lemma~1). (c)~Fresh split D, 7B: mean control latency
$-36\%$, zero output tokens and protocol failures; fp32 replay
(split~A): mean graph deviation ${\approx}0.05$.}
\label{fig:method}
\end{figure*}

\section{Introduction}
\label{sec:intro}

Hierarchical retrieval~\cite{raptor,memwalker,pageindex,lattice} turns
evidence gathering into sequential control: an LLM repeatedly selects $b$
of $k\!\le\!20$ candidate sections and descends into the
selected subtrees. The action space is finite and input-dependent, yet
deployed systems execute this controller by \emph{decoding text}. That
choice imposes a three-way dilemma. Free-form JSON leaves a protocol
surface that dominates the error budget ($3$--$65\%$ of questions fail
on syntax, identifiers, or cardinality in our companion interface
study). Scoring all $k$ verified single-token labels in one
prefill --- the unary shortcut --- removes the protocol surface but also
the controller's memory: selection can no longer condition on earlier
selections. Grammar-constrained decoding~\cite{picard,gcd} keeps
the history-dependent policy and syntactic validity, but pays a serial
critical path of $b$ action tokens plus forced syntax tokens, one model
forward each.

Our central observation is that this dilemma is an artifact of the
execution model, not of the controller. A constrained controller with
horizon $b$ has a \emph{bounded reachable future}: every state it can
visit is a prefix over at most $b$ distinct actions --- for greedy
control at our deployed horizon of three actions, at most $k$ states
remain once the first is fixed. We therefore
\emph{compile} the action grammar into its prefix tree and evaluate all
reachable states at once: one root prefill resolves the first action; one
\emph{tree-causal probe} --- a single forward pass over the concatenated
prefix continuations, with a block-causal mask and serial position
indices --- returns every prefix-conditioned action distribution the
controller can consult. Because each tree state sees exactly the tokens,
positions, and causal ancestry it would see under serial decoding, its
logits \emph{are} the serial controller's logits, and the greedy chain is
recovered arithmetically. We do not approximate or distill the serial
controller; we recompute it on a shallower execution graph
(finite precision can still flip near ties; Sec.~\ref{ssec:equiv}). The critical
path drops from $b+m_{\mathrm{syn}}$ dependent forwards to two,
independent of $b$; no token is ever generated; the protocol surface is
zero by construction.

Compilation also upgrades what the controller can decide. The probe
exposes the controller's \emph{joint} preferences, so unordered
multi-action selection becomes exact posterior inference rather than a
serialization artifact: we marginalize action order and certify the MAP
set with a branch-and-bound certificate --- no truncation heuristics. And the probe compares the two policies: the
AR-vs-unary KL divergence bounds --- via Pinsker --- the
expected-utility loss of the unary shortcut per input.
The construction applies to any finite-action constrained LLM controller
(tool choice, routing, agenda selection); hierarchical RAG is our case
study.

\noindent\textbf{Contributions.}
(1)~\emph{Finite-action control compilation}: constrained controllers
executed as one root prefill plus one tree-causal probe, zero decoded
tokens (Sec.~\ref{ssec:compile}).
(2)~\emph{Policy-exactness theory}: prefix-state equivalence and policy
equivalence under matched tokens, positions, masks, and tie-breaking,
with serial depth exchanged for bounded parallel width
(Sec.~\ref{ssec:theory}).
(3)~\emph{Certified set inference and interaction mass}:
an order-marginalized set posterior with a branch-and-bound MAP
certificate, and a bound on the expected cost of unary
execution (Sec.~\ref{ssec:setmap}).
(4)~\emph{Evaluation}: four document-disjoint splits, two scales, two
accelerators, and a hierarchy-corruption stress test --- policy
fidelity, $100\%$
certification in non-trivial rounds, and $35$--$47\%$ lower control
latency (Sec.~\ref{sec:setup}).

\section{Related work}
\label{sec:related}

\textbf{Hierarchical RAG as control.} Tree- and memory-structured
retrieval~\cite{raptor,memwalker,pageindex,lattice,rdr2} steers an LLM
through document structure; retrieval grounding remains a standard
hallucination mitigation~\cite{ragoriginal}, and evidence quality is
sensitive to structure and OCR noise~\cite{ohrbench,multidocfusion}. A companion
interface study (under review) locates this stack's dominant failures in
the protocol layer; here we change the execution of its strongest
constrained controller.

\noindent\textbf{Constrained decoding and logit selection.} Grammar- and
schema-constrained decoding~\cite{picard,gcd,guideddecoding,jsonschemabench}
enforces validity but keeps the serial path; first-token logit selection
\cite{ma2025logits} avoids decoding but abandons the
sequential policy, inheriting label-position sensitivities
\cite{labelbias,optionorder}. Compilation sits between the two: it
preserves the sequential policy at the cost of one additional forward
pass.

\noindent\textbf{Tree-parallel execution.} Token-tree verification is an
established primitive in speculative and multi-head decoding
\cite{specinfer,medusa}, with IO-aware kernels~\cite{deft}.
Speculation \emph{accelerates} decoding --- draft, verify, reject. Our
compiled tree is not a guess but the \emph{exhaustive} reachable state
space of a bounded-horizon control grammar: nothing to reject, no
decoding left to accelerate, and an equivalence that is provable rather
than distributional.

\noindent\textbf{Parallel structured selection.} Contextual Plackett--Luce learns
pairwise interaction heads for subset selection~\cite{cpl}; mean-field
parallel decoding coordinates token commitments via predictive
interactions~\cite{mfpd}. Both approximate or learn joint structure; we
evaluate the frozen model's exact prefix conditionals without training
and certify the MAP. SimpleTool retrains the function-call representation for parallel
fields~\cite{simpletool}; LLMCompiler parallelizes \emph{external} call
orchestration~\cite{llmcompiler}; neither compiles a frozen
controller's own conditionals, as we do exactly.

\section{Compiled autoregressive control}
\label{sec:method}

\subsection{The serial constrained controller}
\label{ssec:serial}

At a traversal round with candidates bound to contextually verified
single-token labels $\mathcal A_x$ ($|\mathcal A_x|=k$; larger frontiers
are batched deterministically), the grammar $\mathcal G_x$ forces a JSON
array of exactly $n=\min(b,k)$ distinct labels: forced syntax segments
alternate with label slots masked to unused labels. Serial constrained
decoding emits $\sigma=(a_1,\dots,a_n)$ greedily,
\begin{equation}
a_t=\arg\max_{a\in\mathcal A_x\setminus\{a_{<t}\}}
z\bigl(a\mid x\oplus\rho_t\bigr),
\label{eq:greedy}
\end{equation}
with $\rho_t$ the emitted token prefix before slot $t$ and $z$ the
next-token logits. Each of the $n+m_{\mathrm{syn}}$ emitted tokens costs one
dependent forward: critical path
$D_{\mathrm{serial}}=n+m_{\mathrm{syn}}\approx 9$ for $n{=}3$. (Even an
idealized \emph{chunked} engine that fast-forwards known forced spans
still pays one dependent forward per label slot, $D{=}n$; we measure it
as a second baseline in Sec.~\ref{sec:setup}.) Selected
nodes' children form the next frontier; selected sections are packed into
the reader budget $B$ in selection order.

\subsection{Compilation to an action-prefix tree}
\label{ssec:compile}

Once $a_1$ is fixed, the greedy controller can only ever consult
\begin{equation}
N(k,n)=\textstyle\sum_{d=0}^{n-2}\,(k-1)!/(k-1-d)!
\label{eq:width}
\end{equation}
prefix states --- for $n{=}3$, exactly $k$: the trunk $(a_1)$ and the
branches $(a_1,c)$, $c\ne a_1$. Compiled execution is therefore two
steps. \emph{Root prefill}: one pass over the prompt plus the forced
opening syntax yields slot-1 logits and the shared KV cache; $a_1$
resolves by Eq.~\eqref{eq:greedy}. \emph{Tree-causal probe}: the
$N(k,n)$ prefix continuations --- each the exact forced token sequence
serial decoding would emit --- are concatenated into one query of a few
hundred tokens and evaluated in a single forward under a block-causal
mask: every position attends to the cached prompt and its own ancestors
only, with serial position indices (branches may share or replicate the
trunk; attended key--value sets coincide). The last
position of each row is that prefix's action distribution; $a_2,a_3$
follow arithmetically. The critical path is $D_{\mathrm{comp}}=2$
forwards for any $n$, at parallel width $\Theta(N(k,n))$ probe rows:
serial \emph{depth} is exchanged for bounded parallel \emph{width},
and compilation is profitable whenever
\begin{equation}
T_{\mathrm{probe}}\bigl(N(k,n)\bigr)\;<\;
\textstyle\sum_{t=1}^{n+m_{\mathrm{syn}}} T_{\mathrm{step}},
\label{eq:breakeven}
\end{equation}
the regime of every $n\le 3$ configuration we measure at the deployed
frontier width (Sec.~\ref{ssec:bench}).

\subsection{Policy exactness}
\label{ssec:theory}

\textbf{Lemma 1 (prefix-state equivalence).}
\emph{If tree-packed and serial evaluation give a prefix $\rho$ the same
token ids, position ids, ancestor-visible attention set, and parameters,
then all hidden states over $\rho$ coincide at every layer, hence}
$z_{\mathrm{tree}}(\cdot\mid x,\rho)=z_{\mathrm{serial}}(\cdot\mid
x,\rho)$. \emph{Proof:} induction over layers: embeddings agree;
attention at each position reads an identical key/value set by the mask
construction; MLPs and normalizations are deterministic per-position
maps. \hfill$\square$

\noindent\textbf{Theorem 1 (policy equivalence).}
\emph{With identical admissible-label masks, candidate order, and
first-maximum tie-breaking, compiled and serial greedy control select the
same action sequence for every input:}
$\pi_{\mathrm{comp}}(x)=\pi_{\mathrm{serial}}(x)$.
\emph{Proof:} by Lemma~1 the two graphs produce identical logits at
every slot; identical masks and tie-breaking then make each argmax
identical, and induction over slots propagates the shared prefix.
\hfill$\square$

\noindent\textbf{Corollary 1 (margin stability).} \emph{If a slot's decision
margin under either graph exceeds $2\varepsilon_{\max}$, where
$\varepsilon_{\max}$ bounds the cross-graph logit deviation, then the
compiled and serial decisions at that slot coincide.} Margins come at no
extra cost from the probe, so decisions inside the flip window can be
flagged online against a deviation bound calibrated on replays.

Exact arithmetic is the premise; hardware executes the two graphs with
different floating-point reduction orders, so logits deviate by a
data-dependent $\varepsilon$ and near-tie slots can flip. This is a
property of precision, not of compilation: empirically, every observed
divergence sits at a slot whose margin lies within twice the observed
deviation --- the window where Corollary~1 permits flips
(Sec.~\ref{ssec:equiv}) --- so the two executions coincide up to the
precision the deployment already accepts.

\subsection{Certified set inference and interaction mass}
\label{ssec:setmap}

The reader consumes the selection as a \emph{set}; we marginalize order
for the \emph{decision} and serialize the chosen set for packing by its
highest-probability \emph{expanded} ordering --- order enters only
there. The probe
makes the order-marginalized posterior
$P(S\mid x)=\sum_{\sigma\in\mathrm{Perm}(S)}\prod_t
p(\sigma_t\mid x,\sigma_{<t})$ computable \emph{exactly}, and its MAP
certifiable without full expansion: any unexpanded prefix $\rho$ bounds
all completions by $P(\rho)$, giving $L(S)\!\le\!P(S)\!\le\!U(S)$;
expanding ordered pairs by mass until
\begin{equation}
L(S^\star)>\max_{S\neq S^\star}U(S)
\label{eq:cert}
\end{equation}
\emph{certifies} $S^\star$ as the exact MAP, else the remaining pairs are
expanded in one further probe --- the readout is always exact, never
truncated. (Set-MAP adds one probe pass --- a second only when the
certificate defers; split E: $0.44$\,s/question vs.\ $0.37$ compiled,
$0.63$ serial.)
Empirically the certificate fires after the first $24$ of up
to $380$ pairs in every non-trivial round (frontiers with $k\ge 4$;
smaller frontiers admit a single set) across four splits ($527/527$).

Finally, let $P_{\mathrm U}$ be the Plackett--Luce policy induced by
root logits with removal renormalization. The \emph{sequential
interaction mass}
$\mathcal I_{\mathrm{seq}}(x)=\mathrm{KL}(P_{\mathrm{AR}}\,\Vert\,
P_{\mathrm U})$, computable in closed form from the probe, bounds the
expected cost of the unary shortcut for \emph{any} utility $u\in[0,1]$:
$|\mathbb E_{P_{\mathrm{AR}}}u-\mathbb E_{P_{\mathrm U}}u|\le
\sqrt{\mathcal I_{\mathrm{seq}}/2}$ (Pinsker) --- a statement about the
two stochastic policies; greedy agreement is measured directly in
Sec.~\ref{ssec:equiv}. This is a \emph{post-compilation} fidelity
diagnostic: small $\mathcal I_{\mathrm{seq}}$ guarantees the unary
shortcut loses little expected utility; predicting it
\emph{before} compilation is future work. Measured $\mathcal I_{\mathrm{seq}}$ is
heavy-tailed (median $2.6$, p90 $8.3$ nats over the $139$ fully expanded
rounds of split~B). Only $14\%$ of rounds fall below
$\mathcal I_{\mathrm{seq}}{=}0.5$ (expected-utility gap at most $0.5$):
history dependence is pervasive, so the unary shortcut is rarely
provably benign --- exactly where preserving the autoregressive policy
cheaply matters.

\section{Experiments}
\label{sec:setup}

\textbf{Setup.} QASPER~\cite{qasper} with byte-verified evidence spans.
The metric is AllEvidence@$B$: all spans of an annotated evidence group
$\ge 80\%$ covered in the budget-$B$ packed context; error rows count
zero.
Five splits, labeled A--E in collection order, are all accounted for
below. Prompts and interfaces were frozen on development split A (30
docs/90 questions);
B (120 questions) is the development-confirmatory batch of the companion
interface study; C/D are
document-disjoint splits from the official test split (30 docs/90
questions each; C was consumed by the companion study's policy analysis,
so only D and E remain fresh); E (20 docs/60 questions) is
preregistered: frozen \emph{after} the compiler design, identity
hashes recorded before its gold was read. Qwen2.5-7B/3B-Instruct~\cite{qwen25}, bf16,
eager attention, one MetaX C500, offline; latency is per-question
wall-clock over full traversals, with identical configuration
across interfaces and per-run code-hash manifests; document-cluster
bootstrap CIs (10k resamples, seed 20260916).
Baselines: serial grammar decoding, token-by-token as deployed
grammar engines do (the compiled controller's source policy); an
idealized \emph{chunked} serial engine that fast-forwards forced spans
(one forward per slot); unary projection; sequential-label decoding;
exact-$b$ free generation; legacy capped JSON; and embedding-MMR
diversity~\cite{mmr}. A
byte-preserving false-nesting corruption stresses the hierarchy prior,
and the four-arm runs and benchmark are repeated on NVIDIA H800s with
the same runner, prompts, weights, and eager-attention configuration
(one run per device).

\begin{table}[t]
\centering\small
\caption{Policy fidelity and control cost. ``Agree'': exact
action-sequence identity with the serial controller over the full
traversal (divergences are floating-point near-ties,
Sec.~\ref{ssec:equiv}). Latency: mean (median) s/question. Out tok.:
mean output tokens per question, serial / compiled.}
\label{tab:equiv}
\setlength{\tabcolsep}{3.0pt}
\begin{tabular}{lcccc}
\toprule
Split & Agree & Serial lat. & Compiled lat. & Out tok.\\
\midrule
A (90q) & 80/90 & 0.62 (0.50) & 0.40 (0.33) & 11.7 / 0\\
B (120q) & 101/120 & 0.52 (0.46) & 0.32 (0.28) & 10.4 / 0\\
D (90q) & 79/90 & 0.56 (0.48) & 0.36 (0.32) & 11.2 / 0\\
E (60q) & 52/60 & 0.63 (0.50) & 0.37 (0.30) & 12.9 / 0\\
D, 3B & 78/90 & 0.53 (0.44) & 0.28 (0.24) & 11.2 / 0\\
\bottomrule
\end{tabular}
\end{table}

\begin{figure}[t]
\centering
\includegraphics[width=\columnwidth]{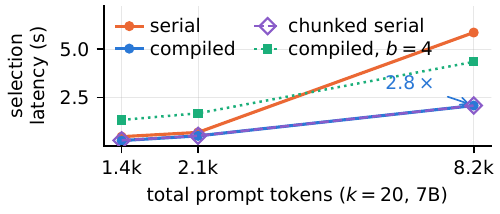}
\caption{Work--depth scaling at the deployed frontier ($k{=}20$, 7B,
synthetic): serial grammar decoding, the idealized chunked engine
($D{=}n$; it overlaps compiled here, while Sec.~\ref{ssec:equiv}
separates them on real workloads), and compiled control ($D{=}2$); the
dotted $b{=}4$ probe shows the measured break-even of
Eq.~\eqref{eq:breakeven}. The 3B model reproduces the pattern.}
\label{fig:results}
\end{figure}

\subsection{Compilation reproduces the serial policy}
\label{ssec:equiv}

Compiled control reproduces the serial controller's exact action
sequences (Table~\ref{tab:equiv}) on $84$--$89\%$ of questions
per split ($87\%$ at 3B) with zero output tokens, cutting
mean control latency by $35$--$41\%$ ($47\%$ at 3B). Selected \emph{sets} agree on
$87$--$90\%$: some divergences merely permute a
selection. Agreement is higher per round ($86$--$92\%$) than per
question: one flipped round redirects the remaining traversal. A
same-run four-arm comparison on D and E (one code hash, arms
interleaved) adds the chunked engine; compiled control beats this
fair baseline too: serial/chunked/compiled run at
$0.58$/$0.37$/$0.35$\,s per question on D and $0.64$/$0.40$/$0.37$ on
E.
% hard column break: page 3 must end at this sentence boundary rather
% than mid-sentence across Fig. 2; remove if any text above reflows.
\newpage
Both cluster-bootstrap CIs on the compiled$-$chunked delta exclude
zero; a reversed arm order on E reproduces the gap ($-0.028$\,s).

The divergences themselves are precision, not policy. Instrumented
replays on identical prefixes show per-slot logit deviations between
the graphs averaging $1.2$ (p95 $2.6$) against serial decision margins
of median $4.1$, and on the margin-instrumented replays (B/D/E) all
$39$ diverging rounds occur at slots with margin $\le 2.5$ --- inside
the $2\varepsilon$ flip window of Corollary~1 --- never at a
confidently separated decision. A full fp32 replay of split A
collapses the deviations by over $20\times$ (mean $0.05$, p95 $0.12$),
and under bf16 the compiled-vs-serial deviation on a slot-level
diagnostic ($1.25$--$2.5$ across slots) is of the same order as
re-chunking the same serial prompt ($2.0$).

Nor are near-tie flips specific to compilation. Chunking alone changes
final selections at least as often ($13$ vs.\ $11$ of $90$ on D; $7$
vs.\ $7$ of $60$ on E), and an NVIDIA H800 repetition preserves the
latency ordering ($0.28$/$0.16$/$0.15$\,s on D; $0.31$/$0.17$/$0.15$
on E) while final-selection agreement drops to $65$--$77\%$ --- the
per-token engine agrees with \emph{itself} across the two
accelerators on just $67$--$73\%$ of questions. The realized bf16
policy is accelerator- and backend-dependent: compiled-versus-serial
disagreement is comparable to the serial policy's cross-accelerator
disagreement.

\subsection{Quality tracks the source policy}
\label{ssec:quality}

Whether history-dependent selection beats unary marginals is
distribution-dependent: the AR policy led unary by $+5.9$/$+4.4$ points
at 2K on splits B/C yet trailed by $1.1$ on D (CI $[-8.9,+6.7]$);
embedding-MMR leaves $84/90$ selections unchanged. Compilation makes
this uncertainty cheap to hedge: both policies come from the same
probe, and $\mathcal I_{\mathrm{seq}}$ (Sec.~\ref{ssec:setmap}) flags
when they can differ. Paired deltas quantify the tracking error:
pooled over four splits and three budgets, compiled minus serial is
$-0.6$ points, CI $[-1.7,+0.5]$ ($1080$ cells);
on the two fresh splits alone, $-2.4$ (CI $[-4.9,-0.4]$) --- a deficit
consistent with bf16 near-tie resolution: fp32 on A restores
$89/90$ identical sequences. Corollary~1 supplies a \emph{parity dial}: rounds
whose minimum margin falls below $\theta$ are replayed serially ---
serial decisions by construction. At $\theta{=}2$, set post hoc above
every observed flip margin, the hybrid's final selections match the
serial engine's on all $150/150$ D$+$E questions ($60\%$ of rounds
gated) --- but at $0.70$--$0.71$\,s it is slower than serial itself,
and on H800 three flips escape the gate ($147/150$): $\theta$, like
$\varepsilon$, is hardware-specific. Exact parity via the dial is not
cheap: compiled trades it for $41\%$ lower latency, and
parity-critical deployments keep the serial engine. Under
false-nesting corruption compiled control keeps a zero protocol
surface ($44.4$ AllEvidence@1K, $0$ errors); free generation fails on
$7.8$--$14.4\%$ of questions.

\subsection{Ablations}
\label{ssec:abl}

On development split A, reading only the controller's \emph{first} label
slot (top-$b$ of slot-1 marginals) collapses mid-budget quality: $42.2$
vs.\ $51.1$ AllEvidence@1K for the full compiled chain --- later,
history-conditioned slots carry information no single readout recovers. Conditionals are \emph{grammar-specific}: a different serialization
of the same task (sequential labels, no JSON syntax) yields a visibly
different policy ($44.4$ vs.\ $50.0$ @1K) --- so we compile the
deployed grammar, not a surrogate prompt. Repeat runs of every
interface are bit-identical on fixed hardware; Table~\ref{tab:equiv}'s
variance comes from the execution graph or precision, never the
controller.

\subsection{Work--depth scaling}
\label{ssec:bench}

Synthetic sweeps ($k\in\{4,\dots,20\}$, $n\in\{2,3,4\}$, prompts
$1.4$k--$8.2$k tokens, both scales) isolate execution cost (Fig.~\ref{fig:results}). At the deployed frontier width $k{=}20$, compiled selection is $1.3$--$2.8\times$ faster than serial decoding for $n\le 3$, and on the MetaX the advantage \emph{grows} with prompt length ($2.8\times$ at 8k tokens; $\ge 2.2\times$ at both scales) because every serial step re-reads the long KV cache; on the H800, whose decode kernels blunt that penalty, gains sit at $1.4$--$2.6\times$, peaking at short prompts. The idealized chunked engine ($D{=}n$) removes most of that per-token penalty ($2.09$\,s at 8k, $n{=}3$, vs.\ $5.85$ serial); compiled control matches it there ($2.09$\,s) and beats it by $6$--$7\%$ on the real workloads of Sec.~\ref{ssec:equiv}, where the remaining dependent forwards still bind. At $n{=}4$ the $362$-row probe loses at short prompts ($0.4$--$0.6\times$ vs.\ token-wise serial), recovers only $1.4\times$ at 8k ($1.1\times$ at 3B), and never catches the chunked engine ($4.34$ vs.\ $2.15$\,s; the wall persists on the H800 at $0.5$--$0.9\times$) --- the width wall that Eq.~\eqref{eq:width} predicts; past it, staged or chunked execution plans win.

Compilation spends
parallel work to shorten the critical path, not to reduce total
computation: at $k{=}20$, $n{=}3$ the probe evaluates $117$ parallel
token positions where serial decoding emits $9$ sequential ones
($13\times$ the work, off the critical path),
and at $n{=}4$ it grows to $362$ rows and $3195$ positions (about
$270\times$) --- exactly why the break-even recedes.
An fp32 probe ($89/90$ parity) runs at $0.55$\,s/question --- no more
than bf16 \emph{serial} decoding ($0.56$\,s); fp32 serial pays
$1.21$\,s; probe memory overhead is $\le 5$\,MB at $n\le 3$ and
${\approx}3.7$\,GB at $n{=}4$.

\section{Conclusion}
\label{sec:concl}

A bounded-horizon constrained LLM controller does not need to be
decoded; it can be compiled. Its reachable future is a shallow prefix
tree that one tree-causal probe evaluates exactly, reproducing the
serial policy in exact arithmetic with a depth-2 critical path and
zero output tokens; certified set inference and an interaction-mass
bound follow. Limits are explicit: probe width
grows factorially beyond the greedy horizon, and bf16 leaves near-tie
divergences that fp32 narrows to $89/90$. The probe is an ordinary attention call,
composable with tree-attention IO optimizations~\cite{deft}; the
guarantees transfer to any finite-action agent loop, and the compiler
and benchmark harness will be released. The controller's semantics
are now a modeling choice decoupled from the serial critical path ---
for hierarchical RAG and every finite-action controller an LLM is
asked to be.

\vfill\pagebreak

% ICASSP 2027 mandatory statements; allowed on page 5 with references.
% Funding confirmed by the corresponding author 2026-09-01 (sole source).
\section{Compliance with ethical standards}
\label{sec:ethics}
This study uses only the publicly available QASPER dataset of scientific
papers and involves no human subjects, animal experiments, or personally
identifiable data; no ethical approval was required.

\section{Funding and conflicts of interest}
\label{sec:funding}
This work was supported by the Shanghai Special Fund for Promoting
High-Quality Industrial Development: Pilot Industry Innovation and
Development Program (Artificial Intelligence Track, Grant
No.~2025-GZL-RGZN-02006). The authors declare no conflicts of interest.

\bibliographystyle{IEEEbib}
\bibliography{refs}

\end{document}